\documentclass[letterpaper, 10 pt, conference]{ieeeconf}  
\IEEEoverridecommandlockouts                              
\usepackage{algorithm}
\usepackage{amsmath,amssymb,amsfonts}
\usepackage{algpseudocode}
\usepackage{array}
\usepackage{epsfig} 
\usepackage[caption=false,font=footnotesize,labelfont=sf,textfont=sf]{subfig}
\usepackage{textcomp}
\usepackage{stfloats}
\usepackage{url}
\usepackage{verbatim}
\usepackage{graphicx}
\usepackage{cite}
\usepackage{kotex}
\usepackage{bm} 

\newtheorem{lemma}{Lemma}

\usepackage{color}
\usepackage{color,soul}
\usepackage[draft]{microtype}
\usepackage{float}
\usepackage[caption=false,font=footnotesize]{subfig} 
\usepackage{adjustbox}
\usepackage{siunitx}
\usepackage{multirow}
\newcommand\ChangeRT[1]{\noalign{\hrule height #1}}

\title{\LARGE \bf

Zero Wrench Control via Wrench Disturbance Observer for Learning-free Peg-in-hole Assembly

}

\author{Kiyoung Choi, Juwon Jeong and Sehoon Oh
\thanks{Kiyoung Choi, Juwon Jeong and Sehoon Oh are with the Department of Robotics and Mechatronics Engineering, DGIST (Daegu Gyeongbuk Institute of Science and Technology), Daegu 42988, Republic of Korea (e-mail: {\tt\footnotesize kychoi@dgist.ac.kr}, and {\tt\footnotesize sehoon@dgist.ac.kr})}
}
         
\begin{document}

\maketitle
\thispagestyle{empty}
\pagestyle{empty}

\begin{abstract}
This paper proposes a Dynamic Wrench Disturbance Observer (DW‑DOB) designed to achieve highly sensitive zero‑wrench control in contact--rich manipulation. By embedding task-space inertia into the observer’s nominal model, DW-DOB cleanly separates intrinsic dynamic reactions from true external wrenches. This preserves sensitivity to small forces/moments while ensuring robust regulation of contact wrenches. A passivity--based analysis further demonstrates that DW-DOB guarantees stable interactions under dynamic conditions, addressing the shortcomings of conventional observers that fail to compensate for inertial effects. Peg‑in‑hole experiments at industrial tolerances (H7/h6) validate the approach, yielding deeper, more compliant insertions with minimal residual wrenches and outperforming a conventional wrench disturbance observer and a PD baseline. These results highlight DW‑DOB as a practical, learning‑free solution for high‑precision zero‑wrench control in contact--rich tasks.
\end{abstract}

\section{Introduction}

With the rise of Physical AI, increasing attention is being directed toward overcoming the inherent limitations of robotic manipulation through advanced control and learning strategies~\cite{miriyev2020skills,sitti2021physical,chen2025advancing}. In this context, zero wrench control (ZWC) is considered a potential enabler for safe and robust interaction, particularly in contact--rich tasks~\cite{villani2016force,hannaford2002time}. Moreover, recent surveys and studies indicate that learning alone is insufficient in contact--rich manipulation unless combined with force/impedance control and passivity-oriented safety layers~\cite{suomalainen2022survey,elguea2023review,beltran2020learning}.

\begin{figure}[t]
\begin{center}
    \includegraphics[width=1\columnwidth]{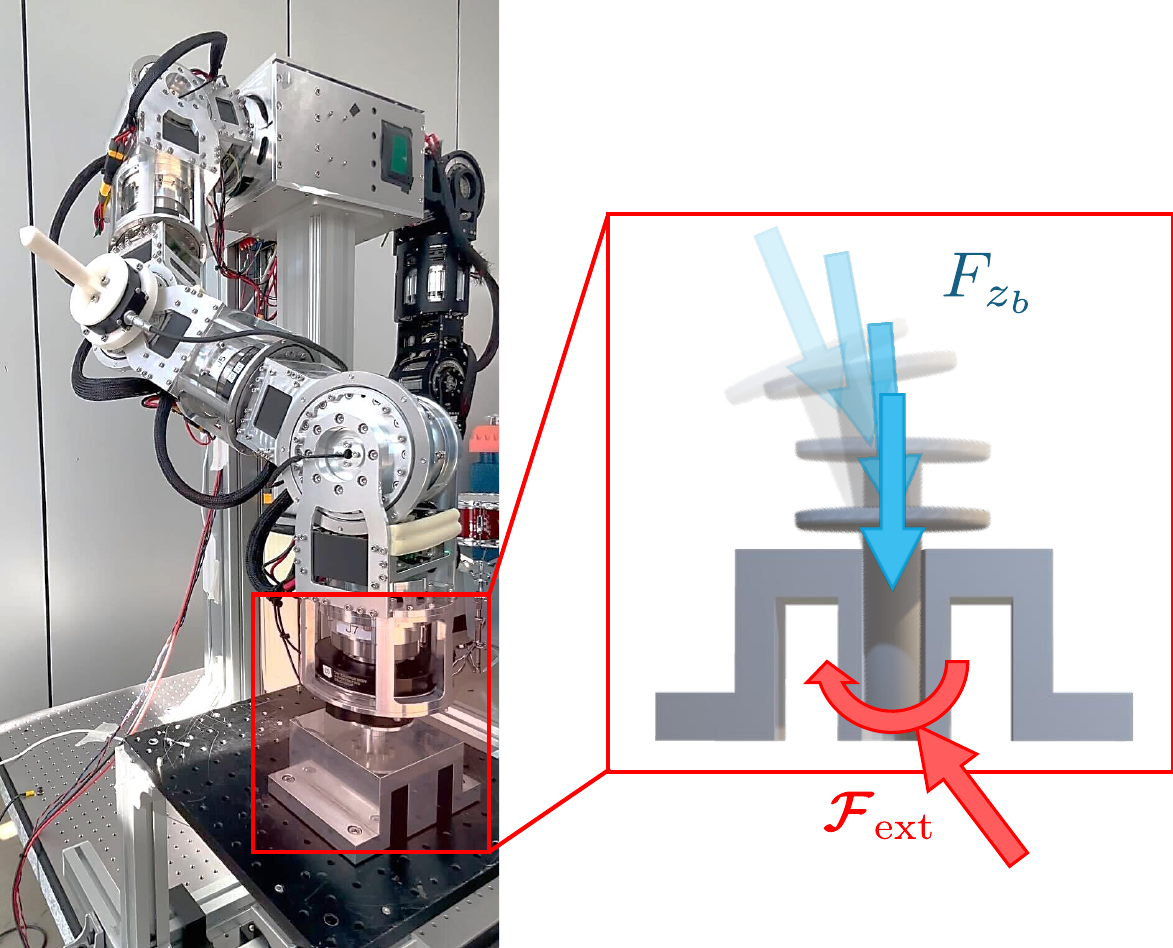}
    \caption{Robot performing a manipulation task that requires highly compliant zero wrench control. The insertion force (blue) in a peg-in-hole task and the resulting external wrench (red) are illustrated.}
    \label{fig:zero wrench control example}
\end{center}
\vspace{-0.5cm}
\end{figure}

Many practical tasks---such as assembling objects, opening doors, or collaborating with humans---inevitably form closed-loop systems that span the ground, robot body, manipulator, object, and ground again. Within such closed-chain interactions, manipulating an object can generate internal stresses that not only compromise task success but may also damage the system or the object itself. Humans naturally mitigate this issue by adjusting their posture and regulating internal wrenches---primarily by releasing wrench components---thereby relieving internal stresses while still achieving high-precision manipulation. A representative example is the peg-in-hole task, in which the manipulator must guide an object into a constrained environment while maintaining compliant contact wrenches throughout the loop. As illustrated in Fig.~\ref{fig:zero wrench control example}, when a manipulator applies an insertion force $F_{z_b}$ along the peg axis (blue), inevitable misalignments produce an external wrench $\bm{\mathcal{F}}_{\mathrm{ext}}$ (red). Without appropriate compliance, these wrenches can generate excessive stresses or jamming at the contact. ZWC enables the manipulator to release 
unnecessary wrench components and be naturally guided by the contact, 
allowing smooth insertion while avoiding damage. In contrast, robotic manipulators still struggle to perform such tasks effectively, largely because they lack the ability to implement human-like zero wrench control. In this paper, zero wrench control (ZWC) is defined as a strategy that regulates the wrench with zero reference, enabling compliant reaction to external wrenches. This capability is crucial for bridging the gap between the robustness of human manipulation and the limitations of current robotic systems.

Among the most widely adopted approaches for enabling compliant interaction in robotic manipulation are impedance control and admittance control. Both methods achieve compliant behavior by rendering a desired virtual dynamics---typically modeled as a mass-damper-spring system---between the robot and the environment. These frameworks have proven effective in guaranteeing stability and compliant interaction under external disturbances in a wide range of tasks. However, both possess inherent structural limitations that restrict their performance in contact--rich manipulation.  

Impedance control is known to be robust in stiff environments, yet its tracking accuracy in free space or under low-impedance conditions often deteriorates due to friction and inner torque-loop effects~\cite{ott2010unified}. On the other hand, admittance control performs well in soft environments but may become unstable or divergent when interacting with stiff environments~\cite{fujiki2022series}. These limitations are further intensified by the bandwidth constraints of internal control loops and by hardware-specific properties, which strongly influence passivity and stability~\cite{focchi2016robot}. Moreover, admittance control is highly sensitive to variations in payload mass or robot inertia, making it vulnerable to performance degradation under parameter uncertainty~\cite{farsoni2017compensation}.  

Industrial manipulators present an additional challenge, as they are typically equipped with fixed inner-loop position controllers that inherently impose high stiffness (low-admittance characteristics). This structural constraint hinders the accurate rendering of the desired admittance and often results in excessive overshoot, large impact forces, and contact instabilities during physical interaction. Consequently, direct implementation of admittance control on industrial manipulators has shown limited effectiveness, especially in tasks requiring high sensitivity such as peg-in-hole insertion.  

Traditional impedance- and admittance-based approaches inherently provide only indirect force regulation by shaping the interaction dynamics. While they can yield compliant behaviors, their reliance on indirect dynamics often prevents them from completely releasing unnecessary internal wrenches during delicate manipulation. In contrast, direct force control explicitly regulates contact wrenches, making it more suitable for tasks where excessive internal stresses or transient impacts must be avoided. In this regard, ZWC is particularly attractive, as it allows a manipulator to relieve internal stresses much like a human operator, thereby achieving both safe and compliant behavior in complex contact--rich tasks.

Recent industrial demands extend well beyond classical peg-in-hole insertion and encompass a wide variety of \textit{contact--rich tasks}, such as connector fastening in electronics assembly, screw tightening and torque regulation in automotive manufacturing, polishing and deburring in aerospace and metal processing, and human--robot collaboration in cooperative manipulation. Despite their diversity, these tasks share a common requirement: precise regulation of force/torque and stable absorption of transient impacts. This underscores the need for control strategies that move beyond indirect force control schemes and toward more \textit{direct and sensitive regulation of contact wrenches}.  

Realizing ZWC in practice requires robust estimation and compensation of unmodeled dynamics and external contact forces. Disturbance observers (DOBs) have been extensively studied in motion control, where they are widely used to suppress the effects of model uncertainties, friction, and external disturbances. Empirical studies have demonstrated that DOBs enhance trajectory-tracking accuracy and improve robustness against disturbances in various motion control problems. More recently, several works have attempted to combine admittance control with DOB to simultaneously improve stability and performance in contact--rich manipulation. For instance, Samuel \textit{et al.} proposed an outer-loop admittance controller 
augmented with DOB for industrial manipulators~\cite{samuel2023task}. Sakaino \textit{et al.} also reported a task-space force-based DOB that estimates disturbances using an identity--nominal model~\cite{sakaino2013force}. However, most existing studies remain joint-space oriented. 
In Cartesian force control, prior DOB designs have been largely limited to quasi--static scenarios with non-zero force/moment references. Studies explicitly addressing dynamic, zero--reference wrench control remain underexplored.

To address these gaps, this paper introduces a novel Cartesian wrench-control framework based on a Dynamic Wrench Disturbance Observer (DW-DOB), specifically designed to implement ZWC. The proposed method is rigorously analyzed and experimentally validated through peg-in-hole experiments, serving as both a representative benchmark and evidence of broader applicability to diverse contact--rich industrial tasks. The main contributions of this paper are as follows:  

\begin{enumerate}
    \item \textbf{Cartesian force-loop DOB with inertia-integrated nominal model:}  
    A nominal model of the disturbance observer that incorporates the inertial dynamics of the robot is designed for high-performance zero wrench control. This design enables robust wrench regulation even under variations in contact--rich environments and during dynamic motions.
    
    \item \textbf{Passivity-based stability analysis:}  
    The proposed DOB-based wrench control is validated both theoretically and experimentally, with a passivity-based stability analysis ensuring robustness under high-impact interactions.
    
    \item \textbf{Learning-free ZWC-based precise peg-in-hole assembly:}  
    By implementing the proposed DW-DOB, the system successfully executes peg-in-hole tasks with industrial tolerance levels (H7/h6) without relying on complex learning-based methods. Experiments confirm successful execution, while suppressing peak insertion forces and reducing potential part damage. 
\end{enumerate}

\vspace{-.2cm}
\section{Problem Formulation} \label{sec:problem}

\subsection{Motivation and Problem Definition}
\subsubsection{Lack of Dynamic Cartesian Wrench Control}
The theoretical foundation for Cartesian force control originates from Khatib’s Operational Space Formulation~\cite{khatib1987unified}, which established a consistent framework for motion and force regulation using the task-space inertia matrix ($\bm{\Lambda}$) and Jacobians. Subsequent work by Chiaverini~\cite{chiaverini2002survey} and Seraji~\cite{seraji1987adaptive} extended these ideas to hybrid position/force control and adaptive hybrid strategies, enabling simultaneous regulation of position and static contact forces. However, these studies were primarily limited to hybrid structures and did not directly contribute to improving the dynamic performance of the Cartesian force loop. A task description that explicitly incorporates the Cartesian wrench---including both force and moment components under dynamic conditions---remains underexplored. Existing frameworks often assume static force equilibrium or orthogonality between motion and force spaces, thereby failing to capture the highly dynamic and sensitive wrench regulation required in contact--rich manipulation.  

\subsubsection{Underdevelopment of DOB for Wrench Control}
Passivity-based methods have been widely adopted to guarantee safe contact in interaction tasks. Hannaford and Ryu introduced Time-Domain Passivity Control (TDPC) to monitor energy exchange during impacts~\cite{hannaford2002time}, while Ortega \textit{et al.} employed energy tanks to maintain passivity under variable impedance settings~\cite{dietrich2016passive,raiola2018development}. These approaches have made significant contributions to stability assurance, but their inherently conservative nature often limits achievable performance.  

In contrast, disturbance observers (DOB) have become powerful tools in motion control, widely applied to suppress disturbances, such as model uncertainties and friction. Chen \textit{et al.} proposed nonlinear DOB (NDO) guaranteeing global stability for manipulators~\cite{chen2000nonlinear}, while Sariyildiz and Ohnishi developed adaptive reaction force observers (RFOB) to improve bandwidth tuning~\cite{sariyildiz2014adaptive}. DOBs have also been extended to force control in series elastic actuator (SEA)-based robots~\cite{sariyildiz2020sliding}. More recently, Samuel \textit{et al.}~\cite{samuel2023task} reported task-space DOB frameworks for trajectory improvement and admittance enhancement. Nevertheless, most of these studies remain joint-space oriented or focused primarily on motion-tracking accuracy; dedicated DOB designs for Cartesian force-loop control, particularly those integrated with passivity considerations for dynamic contact, are still rare.  

\subsubsection{Limitations of Learning-Based Approaches for Peg-in-Hole}
Peg-in-hole (PiH) task has long served as a benchmark for evaluating force-control performance. Classical approaches have relied on compliance-based methods and hybrid force/position strategies, while recent work has explored deep reinforcement learning to acquire generalized insertion strategies~\cite{afifi2022high,shen2025learning}. Although learning-based methods have achieved promising results, they require extensive data collection and computation and still face challenges in generalization. More recent efforts have investigated learning-free frameworks based on compliance control, but few studies have demonstrated reliable success at industrial tolerance levels (e.g., H7/h6).  

Moreover, human-inspired strategies such as ZWC have not been systematically incorporated into robotic control frameworks. As a result, canonical assembly tasks such as peg-in-hole continue to rely heavily on learning approaches rather than on enhancing the underlying force-control architecture. In this work, we show that a DOB-based ZWC framework can achieve industrial-level tolerances without learning, while simultaneously suppressing peak insertion forces to minimize the risk of part damage.  
\vspace{-.2cm}
\subsection{Research Gap and Contributions}
These approaches reveal three key gaps.  
First, dynamic Cartesian wrench control remains insufficiently developed.
Second, DOB theory---though highly effective in motion control---has not been systematically exploited for Cartesian wrench-loop design, especially when integrated with passivity requirements.
Third, control frameworks capable of achieving industrial-tolerance peg-in-hole performance are lacking; instead, most recent advances rely on computationally expensive learning methods.

This paper addresses these gaps by proposing a novel disturbance-observer-based Cartesian wrench-control framework, DW-DOB. The proposed framework enhances the dynamic performance of the Cartesian wrench loop, incorporates passivity-based stability analysis to ensure robustness under impacts, and demonstrates, through peg-in-hole experiments, that industrial-tolerance assembly can be achieved without relying on learning-based methods.

\vspace{-.2cm}
\section{Dynamic Wrench Disturbance Observer (DW-DOB)}
Building on the limitations identified in Sec.~\ref{sec:problem}, we now present a disturbance observer framework tailored for highly sensitive Cartesian wrench control. As discussed in the problem formulation, conventional approaches either oversimplify the force-control loop with static or hybrid formulations, or rely on joint-space DOBs that do not explicitly address Cartesian wrench dynamics. Moreover, conventional force observers often ignore the manipulator’s intrinsic inertia, leading to disturbance estimates corrupted by internal dynamic reactions rather than true external forces. To enable reliable zero wrench control in contact--rich tasks, we therefore propose a \textit{Dynamic Wrench Disturbance Observer (DW-DOB)}, which explicitly incorporates task-space inertia into the disturbance estimation loop. This design ensures accurate separation of external disturbances from the manipulator’s own dynamic responses, even when the system switches between static and dynamic contact conditions.  
\vspace{-.2cm}
\subsection{Dynamics Formulation of Manipulator Wrench}
The task space dynamics of a general $n$-DOF manipulator can be expressed as

\begin{equation}
\bm{\mathcal{F}} = \bm{\Lambda}(\bm{q})\ddot{\bm{x}} + \bm{\mu}(\bm{q},\dot{\bm{q}}) - \bm{\mathcal{F}}_{\mathrm{ext}}, \label{eq:task space dynamics}
\end{equation}
where $\bm{\Lambda}(\bm{q}) = (\bm{J}\bm{M}^{-1}\bm{J}^{\rm T})^{-1}$ is the task-space inertia matrix, $\bm{\mu}(\bm{q},\dot{\bm{q}})$ denotes non-inertial terms including Coriolis/centrifugal and gravity term, and $\bm{\mathcal{F}}_{\mathrm{ext}}$ denotes the external wrench from environmental contact.

The objective of this study is to realize ZWC. To this end, reaction wrenches arising purely from the dynamics of manipulator must be excluded from the control loop.  
\vspace{-.2cm}
\subsection{Contact Wrench Disturbance Observer (CWDOB)}
Conventional joint-space DOBs have been widely used to enhance trajectory-tracking performance or to estimate external forces without dedicated sensors. However, these approaches do not directly regulate the Cartesian force loop itself.  

Task-space force-based DOBs have also been reported, for example by Sakaino \textit{et al.}~\cite{sakaino2013force}, where the nominal model was assumed as the identity matrix. 
We refer to such traditional Cartesian force observers as the Contact Wrench Disturbance Observer (CWDOB). In this case, the estimated disturbance $\hat{\bm{d}}_{CW}$ is given by
\begin{equation}
    \hat{\bm{d}}_{CW}=Q(s)\left( -\bm{\mathcal{F}}_{ext}-\bm{\mathcal{F}}_{c}'\right ), \label{eq:CWDOB_dhat}
\end{equation}
where $\bm{\mathcal{F}}_{\mathrm{ext}}$ is the measured external wrench, $\bm{\mathcal{F}}_{c}'$ is the commanded input force. Here, $Q(s)$ is typically implemented as a first-order low-pass filter of the form  $Q(s)=\omega/(s+\omega)$, where $\omega$ denotes the cutoff frequency. A higher cutoff frequency allows faster disturbance estimation, but it also amplifies sensor noise and increases sensitivity to unmodeled high-frequency dynamics.

While computationally efficient, this formulation neglects manipulator dynamics and thus misinterprets internal inertial reactions as external disturbances. In zero force or zero moment control, such misinterpretation can result in unintended compensation signals and instability, particularly in contact--rich tasks such as peg-in-hole assembly.  
\vspace{-.2cm}
\subsection{Proposed Dynamic Wrench DOB (DW-DOB)}
\begin{figure}[t]
\begin{center}
    \includegraphics[width=1\columnwidth]{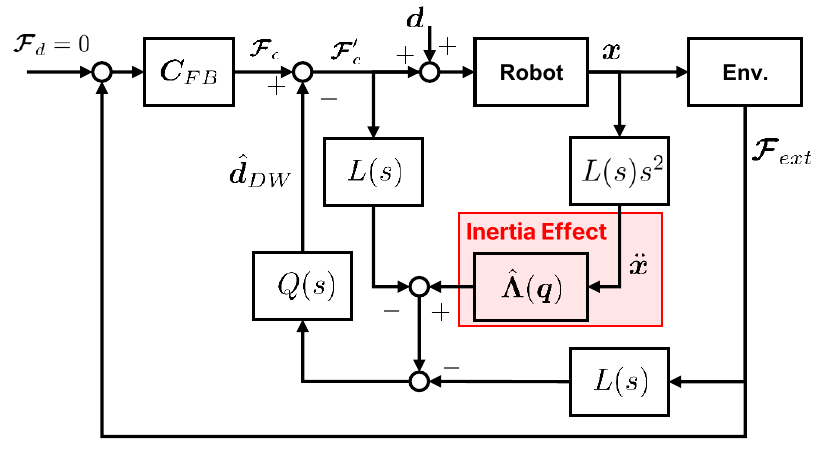}
    \caption{Block diagram of the proposed Dynamic Wrench Disturbance Observer (DW-DOB), which incorporates task-space inertia into the disturbance estimation loop for zero wrench control.}
    \label{fig:DW-DOB diagram}
\end{center}
\vspace{-0.5cm}
\end{figure}

To overcome these limitations, the proposed DW-DOB incorporates the manipulator’s task-space inertia matrix $\bm{\Lambda}$ into the observer. In contact--rich manipulation, the measured wrench $\bm{\mathcal{F}}_{\mathrm{meas}}$ at the end-effector contains both the true external disturbance $\bm{\mathcal{F}}_{\mathrm{ext}}$ and the manipulator’s inertial reaction $\bm{\Lambda}\ddot{\bm{x}}$ to commanded accelerations; conventional observers conflate these terms, misinterpret disturbance estimates and masking of small external forces.  

To explicitly subtract the inertial term, we compute the DW-DOB residual as
\begin{equation}
    \bm{r}_{DW}=\hat{\bm{\Lambda}}(\bm{q}) \ddot{\bm{x}} -\bm{\mathcal{F}}_{ext}-\bm{\mathcal{F}}_{c}', \label{eq:DW-DOB residual}
\end{equation}
where $\ddot{\bm{x}}$ is the task-space acceleration and $\bm{\mathcal{F}}_{c}'$ is the commanded wrench (see Fig.~\ref{fig:DW-DOB diagram}).
In implementation, $\ddot{\bm{x}}$ is obtained from encoder signals by filtered differentiation, specifically, using a first--order filtered differentiator $\mathcal{H}(s)=\omega s/\left( s+\omega \right )$ (possibly cascaded), so the acceleration path realizes a composite filter $L(s)$ with $\ddot{\bm{X}}(s)=L(s)\,s^{2}\bm{X}(s)$.
If the other paths use unfiltered $\bm{\mathcal{F}}_{ext}$ and $\bm{\mathcal{F}}_{c}$ while the acceleration path carries $L(s)$, the phase-lag mismatch in~\eqref{eq:DW-DOB residual} can degrade performance or even destabilize the closed loop in contact--rich settings.
We therefore pre-align phases by applying the same composite filter to the wrench paths and compute the residual as
\begin{equation}
    \tilde{\bm{r}}(s)=\hat{\bm{\Lambda}}(q)\,\ddot{\bm{x}}(s)\;-\;L(s)\,\bm{\mathcal{F}}_{\mathrm{ext}}(s)\;-\;L(s)\,\bm{\mathcal{F}}_{c}(s),
\end{equation}
followed by the disturbance estimate
\begin{equation}
    \hat{\bm{d}}_{DW}(s)=Q(s)\,\tilde{\bm{r}}(s). \label{eq:DW-DOB dhat}
\end{equation}
When $\ddot{\bm{x}}$ is realized by two identical stages, set $L(s)=\mathcal{H}^{2}(s)$; otherwise use the exact composite filter of the acceleration path.

This formulation cleanly separates inertial reactions from true external disturbances, preventing false compensation and improving sensitivity to small contact forces. Consequently, the control loop achieves both compliance and robustness. The manipulator can reject external perturbations while avoiding overreaction to its own dynamics, thereby enabling compliant and robust zero wrench control in highly contact--rich tasks.  

\vspace{-.2cm}
\section{Passivity-based Stability Analysis}
\label{sec:stability}
In this section, we provide a comprehensive analysis of the passivity and stability 
properties of the proposed dynamic wrench disturbance observer (DW-DOB). 
The goal is to highlight the limitations of the CWDOB and demonstrate that DW-DOB ensures robust interaction stability even under dynamic motions. 

\textbf{Assumptions.}
We assume that 
(i) $\bm K_P=\bm K_P^\top$ and $\bm K_D=\bm K_D^\top$ with
\emph{$\mathbf{0}\preceq \bm K_P \preceq \mathbf{I}$} and $\bm K_D\succeq \mathbf{0}$;
(ii) $\bm\mu(q,\dot q)$ and $\dot{\hat{\bm\Lambda}}$ are bounded;
(iii) the uncertainty mismatch $\Delta\bm\Lambda$ is bounded and satisfies
$\|\dot{\bm{x}}^\top\Delta\bm\Lambda\,\ddot{\bm{x}}\|
 \leq c_\Delta\|\Delta\bm\Lambda\|\|\ddot{\bm{x}}\|\|\dot{\bm{x}}\|$;
(iv) the derivative of the external wrench is implemented by a filtered differentiator $\mathcal{H}(s)$.
\begin{equation}
    \dot{\bm{\mathcal{F}}}_{\mathrm{ext}}(s) \;=\; \mathcal H(s)\,\bm{\mathcal F}_{\mathrm{ext}}(s),  
\end{equation}
instead of the raw $\dot{\bm{\mathcal F}}_{\mathrm{ext}}$,
and the filter’s internal energy is added to the storage as
$S_{\mathrm{fil}}=\tfrac{1}{2}\bm z^\top \bm P\,\bm z$ for some $\bm P\succ 0$,
so that $\dot S_{\mathrm{fil}}\le \bm{\mathcal F}_{\!ext}^\top \dot{\bm{x}}
-\kappa(\tau)\|\dot{\bm{x}}\|^2$ with $\kappa(\tau)>0$.

\subsection{System without DOB} \label{subsec:no_dob}

As a baseline controller, we choose PD controller with zero--reference for control input at~\eqref{eq:task space dynamics}
\begin{equation}
    \bm{\mathcal{F}}_c = -\bm{K}_{P}\bm{\mathcal{F}}_{\mathrm{ext}}-\bm{K}_D \dot{\bm{\mathcal{F}}}_{\mathrm{ext}}. \label{eq:PD control input}
\end{equation}
Define the storage function
\begin{equation}
    S = \dfrac{1}{2} \dot{\bm{x}}^\top \hat{\Lambda} \dot{\bm{x}}, 
    \qquad \hat{\bm{\Lambda}} = \bm{\Lambda} + \Delta\bm{\Lambda}.
\end{equation}
Differentiating $S$ yields
\begin{align}
    \dot{S}
    &= \dot{\bm{x}}^\top \hat{\bm{\Lambda}} \ddot{\bm{x}} + \dfrac{1}{2}\dot{\bm{x}}^\top \dot{\hat{\bm{\Lambda}}} \dot{\bm{x}} \nonumber \\
    &= \dot{\bm{x}}^\top(\bm{\mathcal{F}}_c + \bm{\mathcal{F}}_{\mathrm{ext}} - \bm{\mu}) 
       + \dot{\bm{x}}^\top \Delta\bm{\Lambda} \ddot{\bm{x}} 
       + \dfrac{1}{2}\dot{\bm{x}}^\top \dot{\hat{\bm{\Lambda}}} \dot{\bm{x}} \nonumber \\
    &= \bm{\mathcal{F}}_{\mathrm{ext}}^\top \dot{\bm{x}} 
       - \dot{x}^\top \bm{K}_P \bm{\mathcal{F}}_{\mathrm{ext}} - \dot{x}^\top \bm{K}_D \dot{\bm{\mathcal{F}}}_{ext}
       - \dot{\bm{x}}^\top \bm{\mu} \nonumber \\
       &\quad + \dot{\bm{x}}^\top \Delta\bm{\Lambda} \ddot{\bm{x}} 
       + \dfrac{1}{2}\dot{\bm{x}}^\top \dot{\hat{\bm{\Lambda}}} \dot{\bm{x}}.
\end{align}
\begin{lemma}[Cauchy/Young Inequality]
For any $a,b \in \mathbb{R}^n$ and any $\eta>0$, it holds that
\begin{equation}
    a^\top b \;\leq\; \eta \|a\|^2 + \frac{1}{4\eta}\|b\|^2.
\end{equation}
\end{lemma}

Applying Lemma 1 to the cross terms and assuming boundedness of $\bm{\mu}$ 
and $\dot{\hat{\bm{\Lambda}}}$, we obtain reduced margins $\alpha > 0$ and a constant $c>0$ such that
\begin{align}
    \dot{S} &\;\leq\; \bm{\mathcal{F}}_{\mathrm{ext}}^\top \dot{\bm{x}}
    - \alpha \|\dot{\bm{x}}\|^2
    + c\,\|\Delta\bm{\Lambda}\| \|\ddot{\bm{x}}\| \|\dot{\bm{x}}\| \nonumber \\
    &\quad +\dfrac{1}{4\eta_{P}}\| \bm{K}_{P}\bm{\mathcal{F}}_{\mathrm{ext}}\|^{2}+\dfrac{1}{4\eta_{D}}\| \bm{K}_{D}\dot{\bm{\mathcal{F}}}_{ext}\|^{2}+\bm{\Psi}_{PD},
    \label{eq:nodob_passivity}
\end{align}
where $\bm{\Psi}_{PD}$ is bounded constant for $-\dot{\bm{x}}^\top \bm{\mu} + \frac{1}{2}\dot{\bm{x}}^\top \dot{\hat{\bm{\Lambda}}} \dot{\bm{x}}$.

Equation~\eqref{eq:nodob_passivity} shows that the force-feedback PD controller 
introduces two additional non-negative terms, proportional to 
$\|\bm K_P\bm{\mathcal F}_{ext}\|^2$ and 
$\|\bm K_D\dot{\bm{\mathcal F}}_{ext}\|^2$. 
These terms potentially reduce the passivity margin, especially when the external wrench varies rapidly. However under assumptions (i) and (iv), strict output passivity can be preserved.
Thus, the dissipative inequality is restored, ensuring passivity 
preservation for appropriately chosen gains $(\bm K_P,\bm K_D)$ and filter time constant.

\vspace{-.2cm}
\subsection{Contact Wrench Disturbance Observer (CWDOB)}

Next, we analyze the conventional contact wrench disturbance observer (CWDOB). In the presence of a disturbance observer, the total storage function is augmented as
\begin{equation}
    S_{\mathrm{tot}} = S + S_{\mathrm{obs}}, \qquad
    S_{\mathrm{obs}} = \dfrac{1}{2}\bm{\xi}^{\top}\bm{P}\bm{\xi},
\end{equation}
where $S_{\mathrm{obs}}$ accounts for the observer dynamics, and $\xi$ is the observer state. 
The residual for the CWDOB $\bm{r}_{\mathrm{CW}}$ is defined as
\begin{equation}
    \bm{r}_{\mathrm{CW}} := -\bm{\mathcal{F}}_{\mathrm{ext}} - \bm{\mathcal{F}}_c
    ',\quad \bm{\mathcal{F}}_{c}'=\bm{\mathcal{F}}_{c}-\hat{\bm{d}}_{CW}.
\end{equation}
Unlike the DW-DOB, this residual does not include the inertial reaction. Thus, the observer dynamics cannot cancel the inertia term in the storage inequality. By the Q-filter inequality~\cite{willems1972dissipative}, the observer storage derivative becomes
\begin{equation}
    \dot S_{\mathrm{obs}} \;\le\; r_{\mathrm{CW}}^\top \hat{\bm d}
    - \varepsilon \|\hat{\bm d}\|^2, \qquad \varepsilon>0.
\end{equation}
Combining with the baseline derivative and applying Lemma~1 under 
assumptions~(i)--(iv), we obtain
\begin{align}
    \dot{S}_{\mathrm{tot}} &\;\leq\; \bm{\mathcal{F}}_{\mathrm{ext}}^\top \dot{\bm{x}}
    - \alpha' \|\dot{\bm{x}}\|^2 -\varepsilon'\|\hat{\bm{d}}\|^{2}
    + \kappa\,\|\bm{\Lambda}\| \|\ddot{\bm{x}}\| \|\dot{\bm{x}}\| \nonumber \\
    &\quad +\dfrac{1}{4\eta_{P}}\| \bm{K}_{P}\bm{\mathcal{F}}_{\mathrm{ext}}\|^{2}
          +\dfrac{1}{4\eta_{D}}\| \bm{K}_{D}\dot{\bm{\mathcal{F}}}_{ext}\|^{2}
          + \Psi_{CW},
    \label{eq:cwdob_passivity}
\end{align}
where $\alpha'>0$, $\kappa>0$ is a constant, and $\Psi_{CW}$ collects bounded terms 
such as $-\dot{\bm{x}}^\top \bm{\mu} + \tfrac{1}{2}\dot{\bm{x}}^\top \dot{\hat{\bm{\Lambda}}} \dot{\bm{x}}
+ r_{\mathrm{CW}}^\top \hat d - \varepsilon\|\hat d\|^2$.

The key observation is that CWDOB leaves a term proportional to the 
inertia term $\bm{\Lambda}$, i.e., $\kappa \|\bm{\Lambda}\| \|\ddot{\bm{x}}\| \|\dot{\bm{x}}\|$.
This can violate the passivity when accelerations are large. Consequently, CWDOB may appear stable in quasi--static conditions but fails to guarantee robust interaction stability under dynamic motions.
\vspace{-.2cm}

\subsection{Dynamic Wrench Disturbance Observer (DW-DOB)}
\label{subsec:DW-DOB}

We now analyze the proposed dynamic disturbance observer (DW-DOB). 
The key difference from CWDOB is that the observer residual explicitly includes 
the inertial reaction:
\begin{equation}
    \bm{r}_{\mathrm{DW}} := \hat{\bm{\Lambda}} \ddot{\bm{x}} - \bm{\mathcal{F}}_{\mathrm{ext}} - \bm{\mathcal{F}}_c',\quad \bm{\mathcal{F}}_c'=\bm{\mathcal{F}}_c-\hat{\bm{d}}_{DW}.
\end{equation}
Substituting the task-space dynamics~\eqref{eq:task space dynamics} yields
\begin{equation}
    \bm{r}_{\mathrm{DW}} = \Delta\bm{\Lambda} \ddot{\bm{x}} - \bm{\mu}.
\end{equation}
Hence, the exact inertia $\bm{\Lambda} \ddot{\bm{x}}$ is cancelled inside the residual, 
and only the mismatch term $\Delta\bm{\Lambda} \ddot{\bm{x}}$ remains.

Similar to CWDOB, differentiating the storage function and substituting the above definitions gives
\begin{align}
    \dot {S}_{\mathrm{tot}}\le\;&
    \bm{\mathcal{F}}_{\mathrm{ext}}^\top \dot{\bm{x}}
    - \alpha''\|\dot{\bm{\bm{x}}}\|^{2}-\varepsilon'\|\hat{\bm{d}}\|^{2}+\kappa'\|\Delta\bm{\Lambda}\| \|\ddot{\bm{x}}\| \|\dot{\bm{x}}\|\nonumber \\
    &+\dfrac{1}{4\eta_{P}}\| \bm{K}_{P}\bm{\mathcal{F}}_{\mathrm{ext}}\|^{2}
          +\dfrac{1}{4\eta_{D}}\| \bm{K}_{D}\dot{\bm{\mathcal{F}}}_{ext}\|^{2}
          + \Psi_{DW},
\end{align}
where $\alpha''>0$, $\varepsilon'>0$, $\kappa'>0$, and $\Psi_{DW}$ collects 
bounded terms such as $-\dot{\bm{x}}^\top \bm{\mu} 
+ \tfrac{1}{2}\dot{\bm{x}}^\top \dot{\hat{\bm{\Lambda}}} \dot{\bm{x}}$.

Unlike CWDOB, the DW-DOB cancels the full inertial reaction and leaves only the mismatch 
$\Delta\bm{\Lambda}$. As a result, the shortage term is much weaker, and the observer 
dynamics add an additional dissipative term $-\varepsilon' \|\hat {\bm{d}}\|^2$. 
Therefore, DW-DOB preserves port passivity under dynamic motions and ensures robust 
interaction stability even in the presence of model errors.
\vspace{-.2cm}

\subsection{State Stability of CWDOB and DW-DOB}

We adopt a common Lyapunov candidate
\begin{equation}
    V = \dfrac{1}{2}\dot{\bm{x}}^\top \hat{\bm{\Lambda}} \dot{\bm{x}} 
      + \dfrac{1}{2\omega_c}\|\tilde{\bm{d}}\|^2, 
    \qquad \tilde{\bm{d}} = \hat{\bm{d}} - \bm{d}.
\end{equation}
Differentiating $V$ and applying Young’s inequality under Assumptions~(i)--(iv), 
the derivative admits the generic bound
\begin{align}
\dot V \le& 
- c_x \|\dot {\bm{x}}\|^2 - c_d \|\tilde {\bm{d}}\|^2 + \underbrace{\kappa_{\star}\|\bm{\mathcal{I}}_\star\|\,\|\ddot{\bm{x}}\|\,\|\dot {\bm{x}}\|}_{\text{inertia shortage term}} + k_1 \|\bm{d}\|^2 \nonumber \\
&+ k_2+ \dfrac{1}{4\eta_P}\|\bm{K}_P \bm{\mathcal{F}}_{\mathrm{ext}}\|^2
+ \dfrac{1}{4\eta_D}\|\bm{K}_D \dot{\bm{\mathcal{F}}}_{\mathrm{ext}}\|^2,
\label{eq:stability_general}
\end{align}
where $c_x,c_d>0$, $k_i$ are constants, and the inertia shortage term 
depends on the observer:
\[
\kappa_{\star}\|\bm{\mathcal{I}}_\star\| =
\begin{cases}
\kappa\|\bm{\Lambda}\|, & \text{CWDOB (full inertia)} \\
\kappa'\|\Delta\bm{\Lambda}\|, & \text{DW-DOB (uncertainty only)}.
\end{cases}
\]

\noindent
CWDOB leaves a shortage term proportional to the full inertia, which can 
dominate the damping during dynamic motions, leading only to conditional boundedness. 
In contrast, DW-DOB cancels the full inertial reaction, leaving only the mismatch 
term and adding dissipation through $-\|\tilde {\bm{d}}\|^2$, thereby ensuring input-to-state stability (ISS) with respect to bounded inputs $(\bm{\mathcal{F}}_{\mathrm{ext}},\dot {\bm{\mathcal{F}}}_{\mathrm{ext}},\bm{d})$.

\vspace{-.2cm}
\section{Experimental Validation}

\subsection{Experimental Setup}
\begin{figure}[t]
\begin{center}
    \includegraphics[width=0.65\columnwidth]{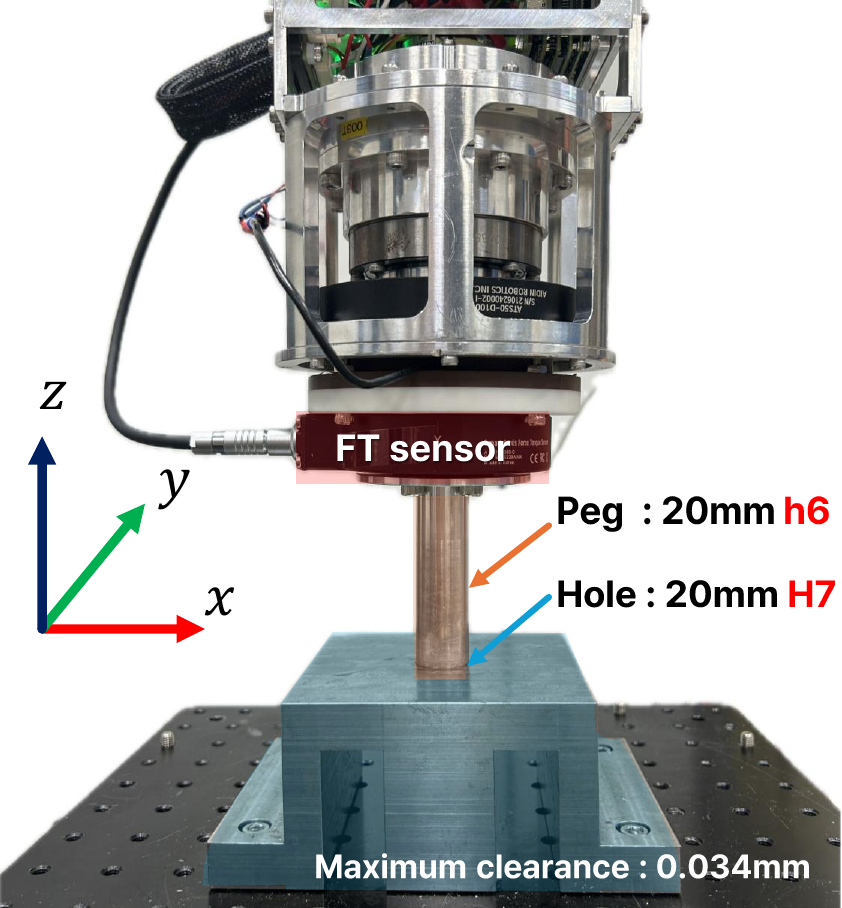}
    \caption{Experiment setup for peg-in-hole task ($\SI{20}{\milli\meter}$ H7/h6).
Maximum clearance $\Delta_{\max}=\SI{0.034}{\milli\meter}$.}
    \label{fig:experiment setup}
\end{center}
\vspace{-0.5cm}
\end{figure}

The proposed DW-DOB was validated experimentally on a 7-DOF robotic manipulator~\cite{deokjin2023ExSLeR} equipped with a force--torque sensor (AIDIN Robotics AFT200-D80-C). The real-time controller was running on real-time Linux (Xenomai 4) at 1 kHz. At each sampling interval, motor current references were sent to the seven joint motors via EtherCAT communication through the motor driver (ELMO Platinum Twitter). The manipulator was tasked with performing a peg-in-hole insertion in the negative z-direction, simulating a delicate assembly task, illustrated in Fig.~\ref{fig:experiment setup}. We assume the hole position is already localized before insertion and that small orientation misalignment remain between peg and the hole. During insertion, the controller regularizes a 4-DOF wrench $\left( F_x, F_y, T_x, T_y \right)$ toward zero while a constant feedforward of $\SI{-20}{\newton}$ is applied along the tool $z_b$-axis to advance the peg. The fit is $\SI{20}{\milli\meter}$ H7/h6 (maximum clearance $\Delta_{\max}=\SI{0.034}{\milli\meter}$), producing a tight, contact--rich condition prone to control instability and stuck without sufficient compliance.

\begin{table}[t]
\caption{Initial pose and controller gains used in the experiments}
\label{tbl:gain_and_pose}
\renewcommand{\arraystretch}{1.4}
\begin{center}
\begin{adjustbox}{width=\columnwidth}
\begin{tabular}{cc !{\vrule width 1.5pt} cccc}
\multicolumn{2}{c!{\vrule width 1.5pt}}{}                                                     & $x$             & $y$             & $w_x$           & $w_y$           \\ \ChangeRT{1.5pt}
\multicolumn{2}{c!{\vrule width 1.5pt}}{$\bm{p}_{init}$}                                           & $-0.216\; \rm{m}$ & $-0.340\; \rm{m}$ & $0.02\; \rm{rad}$ & $0.02\; \rm{rad}$ \\ \hline \hline
\multicolumn{1}{c|}{\multirow{2}{*}{Gain set A}} & $\bm{K}_P$ & 0.10             & 0.10             & 0.50             & 0.50             \\ \cline{2-6} 
\multicolumn{1}{c|}{}                                        & $\bm{K}_D$ & 0.01            & 0.01            & 0.06            & 0.06            \\ \hline
\multicolumn{1}{c|}{\multirow{2}{*}{Gain set B}} & $\bm{K}_P$ & 1.00               & 1.00               & 5.00               & 5.00               \\ \cline{2-6} 
\multicolumn{1}{c|}{}                                        & $\bm{K}_D$ & 0.01            & 0.01            & 0.06            & 0.06           
\end{tabular}\vspace{-.4cm}
\end{adjustbox}
\end{center}\vspace{-.2cm}
\end{table}

The experiment compared three controllers: the proposed DW-DOB, a conventional contact-wrench disturbance observer (CWDOB), and a PD baseline without a disturbance observer. DW-DOB and CWDOB estimate the disturbance according to \eqref{eq:DW-DOB dhat} and \eqref{eq:CWDOB_dhat}, respectively, while all controllers use the same wrench PD structure in~\eqref{eq:PD control input}. For the PD baseline, two gain sets are considered: one with the same gains as the inner PD loop of DW-DOB ($\mathrm{PD}_{\ell}$) and one with higher gains ($\mathrm{PD}_h$). All gains are shown in Table~\ref{tbl:gain_and_pose}. Only $\mathrm{PD}_h$ uses gain set B for the PD gain, whereas the other controllers use gain set A. For DW-DOB, the task-space inertia matrix $\hat{\bm{\Lambda}}$ is computed with the Pinocchio library~\cite{carpentier2019pinocchio} from the robot CAD model. The Q-filter is implemented with a cutoff frequency of \SI{15}{\hertz} (i.e., $\omega=2\pi\cdot 15~\mathrm{rad/s}$), and composite filter $L(s)$ is implemented as a cascade of two first--order low--pass filters with cutoff frequencies $\SI{100}{\hertz}$ and $\SI{15}{\hertz}$. For safety, an emergency stop is triggered if the measured force and moment norms exceed $\SI{90}{\newton}$ and $\SI[inter-unit-product=\ensuremath{\cdot}]{5}{\newton\metre}$, respectively.

Two experimental protocols are conducted. First, all controllers start from the same initial pose to assess the passivity-based analysis and compare insertion performance. Second, DW-DOB performs 15 trials from distinct initial angular misalignments to evaluate robustness and self-alignment capability.

\vspace{-.2cm}

\subsection{Experiment Results and Discussion}
\begin{figure}[t]
	\begin{center}
		\subfloat[]{\includegraphics[height=0.49\columnwidth]{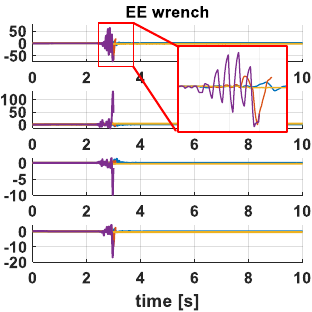}}
		\subfloat[]{\includegraphics[height=0.49\columnwidth]{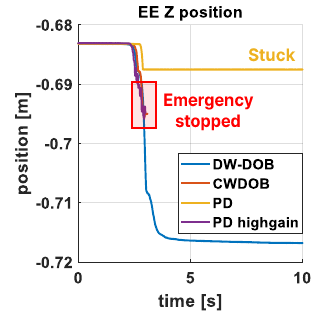} }
		
		\caption{End-effector (a) wrench $(F_x, F_y, T_x, T_y)$ (b) and z-position during insertion, illustrating wrench suppression achieved by DW-DOB compared to the PD controller and CWDOB.}
		\label{fig:exp_result1_wrench_position}
	\end{center}
 \vspace{-0.4cm}
\end{figure}

\begin{figure}[t]
	\begin{center}
		\subfloat[]{\includegraphics[width=0.49\columnwidth]{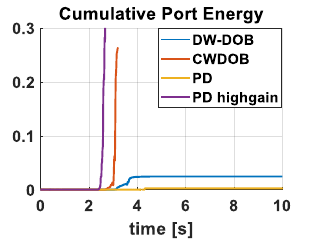}}
		\subfloat[]{\includegraphics[width=0.49\columnwidth]{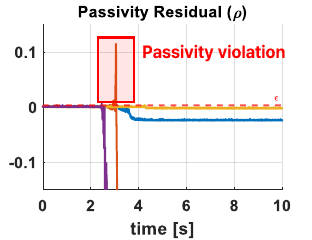} }
		
		\caption{(a) Cumulative port energy $E_{\mathrm{port}}(t)$ and (b) Passivity residual $\rho(t)$ during the peg-in-hole insertion with respect to each controllers.}
		\label{fig:exp result1_passivity}
	\end{center}
 \vspace{-0.4cm}
\end{figure}

For the first experiment, the same initial pose $\bm{p}_{\mathrm{init}}$ was used for all (see Table~\ref{tbl:gain_and_pose}). As shown in Fig.~\ref{fig:exp_result1_wrench_position}, the results indicate that DW-DOB maintains a low residual wrench and achieves the deepest and most consistent $z_b$-insertion. The $\mathrm{PD}_\ell$ case inserts slightly and then sticks because it cannot generate sufficient force to overcome friction and other disturbances, whereas CWDOB and $\mathrm{PD}_h$ produce large contact wrenches and trigger an emergency stop due to unmodeled inertial effects during dynamic contact. In addition, we swept multiple gain settings for the PD baseline beyond $\mathrm{PD}_\ell$ and $\mathrm{PD}_h$; across all tested gains the PD controller either stuck or diverged, triggering the emergency stop and exceeding the safety thresholds of the setup. These observations demonstrate the effectiveness of DW-DOB for compliant zero wrench control in challenging, contact--rich environments.

To empirically verify the passivity claims, we monitored two energy--based metrics at the robot--environment port. First, the cumulative port energy $E_{\mathrm{port}}(t)=\int_{0}^{t}\bm{\mathcal{F}}_{\mathrm{ext}}^{\!\top}\dot{\bm{x}}(\tau)\,d\tau$ integrates the interaction power. With zero initial storage (including the filter storage defined in our passivity analysis), a passive closed loop cannot generate net energy. Accordingly, it must remain bounded and should not increase once sustained contact is established. Second, the passivity residual $\rho = \Delta S-E_{\mathrm{port}}$ quantifies the energy margin relative to the storage $S$: $\rho \le 0$ indicates that the closed loop has not injected net energy, whereas $\rho >0$ signals a passivity violation. For CWDOB, the storage derivative retains a shortage term proportional to the full inertia, $\kappa\|\bm{\Lambda}\|\,\|\ddot{\bm{x}}\|\,\|\dot{\bm{x}}\|$, which can dominate damping during dynamic motion and violate passivity margins---explaining the spikes in the passivity residual and the rapid growth $E_{\mathrm{port}}$ observed in Fig.~\ref{fig:exp result1_passivity}.

\begin{figure}[t]
	\begin{center}
		\subfloat[]{\includegraphics[height=0.49\columnwidth]{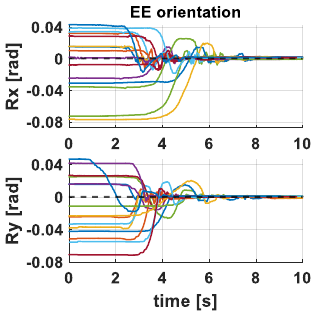}}
		\subfloat[]{\includegraphics[height=0.49\columnwidth]{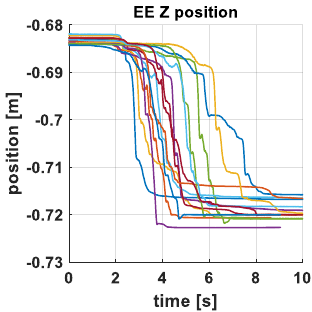} }
		
		\caption{End-effector (a) orientation and (b) $z$-position profiles during 15 insertion.}
		\label{fig:exp_result2_IIWDOB_15times}
	\end{center}
\end{figure}

The second experiment evaluates insertion from 15 different initial angular misalignment. We swept PD gains beyond $\mathrm{PD}_\ell/\mathrm{PD}_h$ under the same safety thresholds; all PD settings and CWDOB either stalled or diverged, triggering the emergency stop. Thus, only DW-DOB admits a 15--trial robustness evaluation. As shown in Fig.~\ref{fig:exp_result2_IIWDOB_15times}, DW-DOB succeeds in all trials, demonstrating strong robustness to initial-pose uncertainty and a self-alignment capability: by releasing unnecessary wrench components, the manipulator is naturally guided into the correct alignment for insertion. The consistent success across all 15 trials underscores the reliability of DW-DOB---crucial for industrial deployment and for human--robot collaboration where predictable responses and adaptability to human--induced variability are essential.

\vspace{-.2cm}

\section{Conclusion}

This paper introduced a novel Cartesian wrench control framework based on a Dynamic Wrench Disturbance Observer (DW-DOB), specifically designed to realize zero wrench control in contact--rich manipulation. By explicitly incorporating the task space inertia into the observer model, the proposed method effectively separates true external wrenches from dynamic reactions, thereby enhancing both sensitivity and robustness. A passivity--based stability analysis established that DW-DOB preserves stable interaction even under dynamic conditions, addressing the limitations of conventional disturbance observers that leave inertia-induced shortage terms.

The effectiveness of the proposed framework was validated through peg-in-hole experiments with industrial tolerances (H7/h6). The results demonstrated that DW-DOB achieves compliant insertion, suppresses peak contact forces, and reduces the risk of jamming or part damage, outperforming both CWDOB and PD baseline. These findings highlight DW-DOB as a practical, learning-free solution for precise force regulation in industrial assembly tasks.

Future work will extend the proposed approach to a broader set of contact--rich applications, including multi-arm cooperation or human--robot collaboration, and investigate integration with adaptive or learning--based strategies to further enhance performance under highly variable environments.



\bibliography{ref}{}
\bibliographystyle{IEEEtran}

\end{document}